\documentclass{dynafront2025} 

\usepackage{multirow}

\title[Distributed Low-Communication Training with Decoupled Momentum Optimization]{Distributed Low-Communication Training with Decoupled Momentum Optimization}

\optauthor{%
\Name{Sasho Nedelkoski} \Email{sasho.nedelkoski@campus.tu-berlin.de}\\
\Name{Alexander Acker} \Email{alexander.acker@logsight.ai}\\
\Name{Odej Kao} \Email{odej.kao@tu-berlin.de}\\
\Name{Soeren Becker} \Email{soeren.becker@logsight.ai}\\
\Name{Dominik Scheinert} \Email{dominik.scheinert@logsight.ai}\\
}

\begin{document}

\maketitle

\begin{abstract}%
The training of large models demands substantial computational resources, typically available only in data centers with high-bandwidth interconnects.
However, reducing the reliance on high-bandwidth interconnects between nodes enables the use of distributed compute resources as an alternative to centralized data center training.
Building on recent advances in distributed model training, we propose an approach that further reduces communication by combining infrequent synchronizations across distributed model replicas with gradient momentum compression.
In particular, we treat the optimizer momentum as a signal and decompose the Nesterov momentum into high- and low-frequency components via the discrete cosine transform (DCT).
Only the high-frequency components are synchronized across model replicas every $H$ steps.
Empirically, our method achieves up to a $16\times$ reduction in communication compared to the baseline DiLoCo, and it generalizes across architectures, including transformer-based language models and convolutional neural networks for images.
Overall, this work advances the feasibility of training large models on distributed nodes with low-bandwidth interconnects.

\end{abstract}


\section{Introduction}
The training of large models requires enormous computational resources, typically distributed across many accelerator nodes connected by highly optimized networking infrastructure~\cite{touvron2023llamaopenefficientfoundation, bai2023qwen, liu2024deepseek, black2022gptneox20bopensourceautoregressivelanguage, penedo2023refinedwebdatasetfalconllm}. 
While distributing training across multiple nodes shortens wall-clock time, it also incurs substantial communication overhead: parameters must be synchronized frequently over specialized networks. 
This requirement constrains the use of heterogeneous or geographically distributed compute resources~\cite{shoeybi2020megatronlmtrainingmultibillionparameter, zhang2024poplarefficientscalingdistributed}.

Two paradigms have been explored to relax this constraint: (1) reducing the synchronization frequency between distributed model replicas and (2) reducing the volume of data exchanged during each synchronization. 
The first approach, rooted in federated learning, performs multiple local model updates before synchronizing model weights or gradients across all or a subset of replicas. 
FedAvg~\cite{McMahanMRA16} and its extension FedOpt~\cite{reddi2021adaptivefederatedoptimization} are foundational methods in this paradigm, aggregating weights and optimizer states globally after $H$ local steps~\cite{karimireddy2021scaffoldstochasticcontrolledaveraging, hsu2019measuringeffectsnonidenticaldata}. 
DiLoCo~\cite{douillard2024dilocodistributedlowcommunicationtraining} builds upon this approach by combining local AdamW optimization with infrequent global Nesterov momentum updates, achieving near-optimal training performance with fewer synchronizations.

Model sparsification represents one method in the second paradigm; however, existing approaches are often limited to specific model architectures~\cite{beton2025improving}. 
A more general strategy for reducing data volume during model replica synchronization is compression. 
Quantization~\cite{pmlr-v202-wang23t} of model weights, activations, and optimizer states is a well-explored technique within this category. 
More recently, methods from signal processing have been applied to selectively synchronize gradient components that carry the most information~\cite{peng2024demodecoupledmomentumoptimization}.


In this work, we combine the two paradigms, focusing on recent advances in applying signal processing for gradient compression and on the integration of regular local AdamW optimization with infrequent global Nesterov momentum optimization.
Specifically, we distribute a model across multiple nodes and perform $H$ local training steps with AdamW. 
During the subsequent global synchronization step, we decompose the local optimizer momentum parameters into frequency components using the discrete cosine transform (DCT) and synchronize only the top-$k$ high-frequency components across all model replicas. 
Each worker then reconstructs the momentum by combining the synchronized high-frequency components with its local low-frequency components.

We evaluate our approach on both transformer (GPT-NeoX) and CNN (ResNet) architectures using the C4 and ImageNet-1k datasets, respectively. 
For the transformer model, our method achieves a $3000\times$ communication reduction compared to DDP with a $6\%$ perplexity increase, and a $16\times$ reduction compared to DiLoCo $H=128$ at $2.5\%$ perplexity increase, with similar trends observed for the CNN model on ImageNet-1k.



\section{Method}
We define the training dataset as a finite set of input–output pairs
$D = \{(x_i, y_i)\}_{i=1}^N, \quad x_i \in \mathcal{X}, \; y_i \in \mathcal{Y}$
where $\mathcal{X}$ denotes the input space, $\mathcal{Y}$ the output space, and $N$ the number of examples. Our focus is on the data-parallel setting, where the dataset is randomly partitioned across a set of worker nodes, each training a distinct model replica.
Our objective is to optimize the model training for a low-bandwidth network environment, where the synchronization between replicas is the primary bottleneck.

\begin{algorithm2e}[t]
\caption{Distributed Low-Communication Training with Momentum Decomposition}
\label{alg:diloco_demo}
\SetKwInOut{Require}{Require}

\Require {Initial model $\theta^{(0)}$, $W$ workers, optimizers InnerOpt and OuterOpt, and learning rates $\eta_{\text{inner}}, \eta_{\text{outer}}$, momentum decay $\beta \in (0,1)$, mixing coefficient $\alpha \in [0,1]$, top-$k$ components, momentum states $m_w^{t-1}$ with $m_w^0 = 0$}

\For{$t \gets 1$ \KwTo $T$}{%
    \For{$w \gets 1$ \KwTo $W$}{%
        $\theta_k^{(t)} \gets \theta^{(t-1)}$\;
        \For{$h \gets 1$ \KwTo $H$}{%
            
            $\theta_w^{(t)} \gets \textsc{InnerOpt}(\theta_w^{(t)}, \nabla L)$\;%
            
        }
        \textsc{OuterOpt}: \\
        $g_w \gets \theta_k^{(t)} - \theta^{(t-1)}$ \\
        $m_w^t \gets \beta m_w^{t-1} + g_w$ \\
        $q_w \gets \textsc{ExtractHighFreqComponentsWithDCT}(m_w^t, k)$ \\
        $m_w^t \gets m_w^t - \textsc{InverseDCT} (q_w)$ \\
        $Q_t \gets \textsc{InverseDCT}(\textsc{Synchronize}(q_w))$ \\
        $m_w^t \gets m_k^t + \alpha Q_t$ \\
        $g_w \gets \alpha g_k + \alpha\beta m_w^t + (1 - \alpha) Q_t$ \\
        $\theta_w^{(t)} \gets \theta^{(t-1)} - \eta_{\text{outer}} g_w$\;%
    }
}
\end{algorithm2e}

\subsection{Distributed Low-Communication Training with Momentum Decomposition}

Our method builds upon the paradigm introduced by federated learning and recently explored by DiLoCo~\cite{douillard2024dilocodistributedlowcommunicationtraining}, shown in \ref{appendix:diloco}.
Thereby, communication overhead is reduced by training each model replica locally over multiple steps using AdamW applying a global synchronization.
DiLoCo uses cumulative gradients locally and executes momentum-based aggregation using an outer optimizer during the global synchronization. 
To further reduce the data volume that each node sends and receives, we implement DCT to decompose the momentum vectors into frequency components. 
Only the high-frequency components of the momentum are synchronized. 
Low-frequency components are kept in a local momentum aggregate to accumulate over time. 
A similar approach is explored in DeMo~\cite{peng2024demodecoupledmomentumoptimization} (see ~\ref{appendix:demo}), where it is shown that synchronizing momentum's high-frequency components reduces communication volume while preserving convergence properties.

We detail our proposed method in Algorithm~\ref{alg:diloco_demo}. 
Beginning with the outer iteration, each worker initializes from the last global model $\theta^{(t-1)}$ and performs $H$ steps of local updates on its data shard. 
Afterwards, each worker computes a pseudo-gradient, $g_w = \theta_w^{(t)} - \theta^{(t-1)}$, representing the cumulative local update.
This pseudo-gradient is then used to update the worker’s local momentum state $m_w^t = \beta m_w^{t-1} + g_w$.

Next, we apply the momentum decomposition with DCT to extract the top-$k$ high-frequency components. 
Intuitively, the high-frequency components represent the most important, rapidly changing directions in the gradient. 
The key hypothesis is that sharing these high-frequency components more frequently will align all workers on the most critical updates, thereby preserving convergence~\cite{peng2024demodecoupledmomentumoptimization}.

After each worker computes its own $q_w$, temporary it reconstructs $q_w$ with inverse DCT and subtracts from the momentum, leaving $m_w^t$ with only the remaining low-frequency components to continue accumulating locally. 
This ensures no significant information is permanently ignored. 
Over multiple iterations, these low-frequency components can accumulate and eventually gain enough magnitude to be captured in the high-frequency set, progressively integrating their information into the global model. 
This process can be viewed similar to error-feedback SGD with momentum~\cite{karimireddy2019errorfeedbackfixessignsgd}, where storing and feeding back residual errors is key for convergence.

Next, $q_w$ are synchronized accross all workers using all-gather operation. After synchronization, each worker performs inverse DCT, now on the synchronized high-frequency components. These are then added to the momentum, completing the decoupling momentum step.
Empirically, we observed that controlling the influence of high-frequency components is crucial, with the optimal strategy depending on the dataset and model architecture. 
To this end, we introduce a parameter $\alpha$ to modulate their contribution. 
When $\alpha=1$, synchronized high-frequency components are fully integrated into the local momentum, resembling a standard momentum update. 
When $\alpha=0$, the local momentum accumulates only low-frequency components, while high-frequency components are handled separately—similar to DeMo~\cite{peng2024demodecoupledmomentumoptimization} in the DDP setting. 
The optimal $\alpha$ varies with the dataset and architecture. 
While a formal convergence proof is left for future work, our rationale is further supported by the intuitive explanation in~\ref{appendix:interaction}.

\section{Results}
We evaluate our distributed training method on two model architectures: (1) a transformer-based language model (GPT-NeoX)\cite{black2022gptneox20bopensourceautoregressivelanguage} trained on the C4 dataset\cite{raffel2023exploringlimitstransferlearning}, and (2) a convolutional neural network (ResNet)\cite{he2015deepresiduallearningimage} trained on ImageNet-1k\cite{5206848}. 
During training, we measure the perplexity on the respective test sets and record the total cumulative communication volume - i.e. sent and received bytes - aggregated across all worker nodes. 
Further details of the experimental setup are provided in \ref{appendix:experimental_setup}.

\begin{figure}[ht]
  \centering
  \begin{minipage}[b]{0.43\linewidth}
    \centering
    \includegraphics[width=\linewidth]{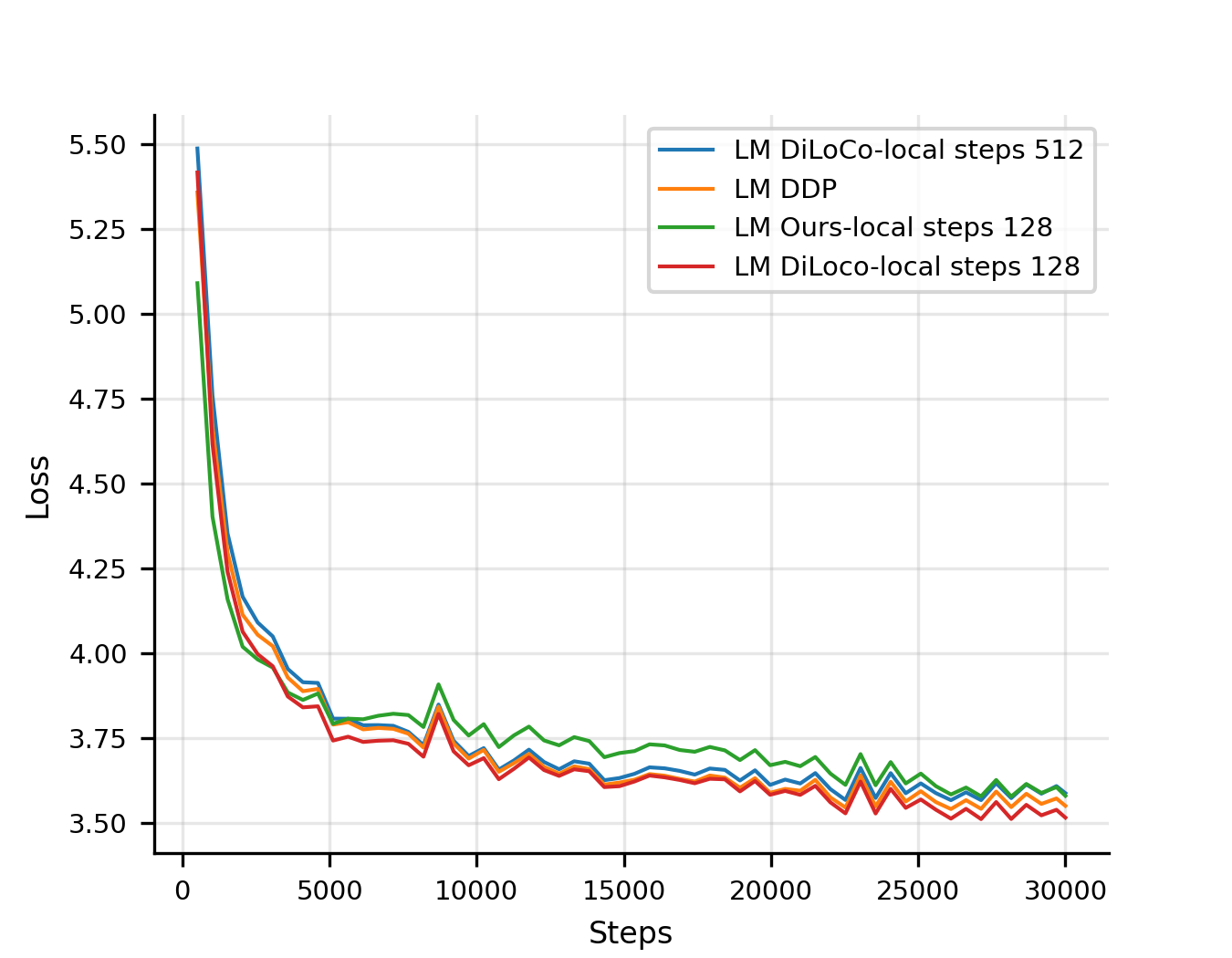}
    \\[2pt]
    \label{fig:demoloco_lm_loss}
  \end{minipage}\hfill
  \begin{minipage}[b]{0.43\linewidth}
    \centering
    \includegraphics[width=\linewidth]{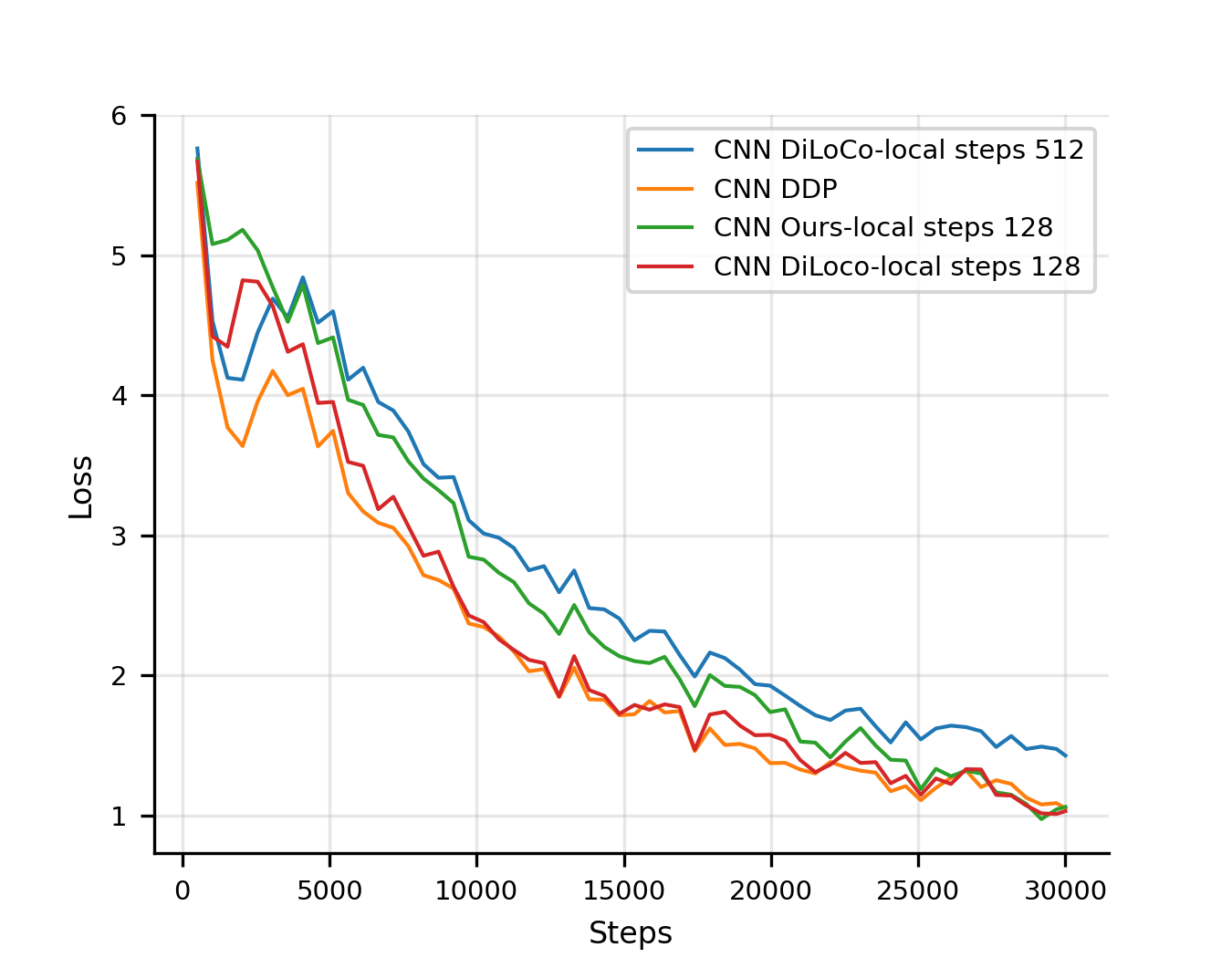}
    \\[2pt]
    \label{fig:demoloco_cnn_loss}
  \end{minipage}
  \caption{(a) Language modeling training loss; (b) Image classification training loss}
  \label{fig:combined_loss}
\end{figure}

\vspace{-0.5em}
\begin{table*}[h]
\centering
\resizebox{0.85\textwidth}{!}{%
\begin{tabular}{l|cccccccc|}
\cline{2-9}
 & \multicolumn{4}{c|}{GPT-Neo-X (C4)} & \multicolumn{4}{c|}{ResNet (ImageNet-1k)}   \\ \hline
\multicolumn{1}{|l|}{Methods}   & \multicolumn{1}{c|}{\begin{tabular}[c]{@{}c@{}}DDP\\ $H$=$1$\end{tabular}} & \multicolumn{1}{c|}{\begin{tabular}[c]{@{}c@{}}DiLoCo\\ $H$=$128$\end{tabular}} & \multicolumn{1}{c|}{\begin{tabular}[c]{@{}c@{}}DiLoCo\\ $H$=$512$\end{tabular}} & \multicolumn{1}{c|}{\begin{tabular}[c]{@{}c@{}}Ours\\ $H$=$128$\end{tabular}} & \multicolumn{1}{c|}{\begin{tabular}[c]{@{}c@{}}DDP\\ $H$=$1$\end{tabular}} & \multicolumn{1}{c|}{\begin{tabular}[c]{@{}c@{}}DiLoCo\\ $H$=$128$\end{tabular}} & \multicolumn{1}{c|}{\begin{tabular}[c]{@{}c@{}}DiLoCo\\ $H$=$512$\end{tabular}} & \begin{tabular}[c]{@{}c@{}}Ours\\ $H$=$128$\end{tabular} \\ \hline
& \multicolumn{8}{c|}{2 Worker Nodes} \\ \hline
\multicolumn{1}{|l|}{Perpl.}    & \multicolumn{1}{c|}{$34.13$} & \multicolumn{1}{c|}{$35.35$} & \multicolumn{1}{c|}{$36.37$} & \multicolumn{1}{c|}{$35.48$} & \multicolumn{1}{c|}{$6.54$} & \multicolumn{1}{c|}{$5.21$} & \multicolumn{1}{c|}{$7.68$} & $7.49$ \\ \hline
\multicolumn{1}{|l|}{Com.(GB)} & \multicolumn{1}{c|}{$20\times10^4$}  & \multicolumn{1}{c|}{$94.0$} & \multicolumn{1}{c|}{$23.6$} & \multicolumn{1}{c|}{$6.7$} & \multicolumn{1}{c|}{$5\times10^3$} & \multicolumn{1}{c|}{$22.9$} & \multicolumn{1}{c|}{$5.7$} & $1.9$ \\ \hline
& \multicolumn{8}{c|}{4 Worker Nodes}  \\ \hline \multicolumn{1}{|l|}{Perpl.} & \multicolumn{1}{c|}{$34.12$} & \multicolumn{1}{c|}{$33.33$} & \multicolumn{1}{c|}{$35.99$} & \multicolumn{1}{c|}{$35.48$} & \multicolumn{1}{c|}{$9.69$} & \multicolumn{1}{c|}{$7.26$} & \multicolumn{1}{c|}{$8.84$} & $9.02$ \\ \hline
\multicolumn{1}{|l|}{Com.(GB)} & \multicolumn{1}{c|}{$30 \times 10^4$} & \multicolumn{1}{c|}{$141.0$} & \multicolumn{1}{c|}{$35.5$} & \multicolumn{1}{c|}{$16.8$} & \multicolumn{1}{c|}{$7.5\times10^3$} & \multicolumn{1}{c|}{$34.3$} & \multicolumn{1}{c|}{$8.6$} & $4.6$ \\ \hline
\end{tabular}
}
\caption{Validation perplexity and communication in gigabyte with two and four nodes.}
\label{tab:perplexity_comm_nodes}
\end{table*}

\subsection{Evaluation loss and communication}
We compare our method against a standard DDP setup and DiLoCo with synchronization intervals $H \in {128, 512}$, evaluating training loss, validation perplexity, and communication volume across varying model architectures and node scales.

Figure~\ref{fig:combined_loss} shows training loss trends for the different model architectures. 
Our approach exhibits a slower initial stabilization phase, similar to DiLoCo-512, due to its decoupled momentum updates and heavy compression. 
This delayed convergence mirrors error-feedback mechanisms, requiring a warm-up period to accumulate momentum components before effective synchronization. 
However, in later stages, our method converges faster than baselines. 
Similar patterns appear in CNN-based image classification tasks.

Table~\ref{tab:perplexity_comm_nodes} summarizes validation perplexity and communication costs on GPT-Neo-X (C4) and ImageNet-1k (ResNet) across 2 and 4 nodes. 
Our method with $k=32$ consistently matches or outperforms baselines with reduced communication. 
On C4, it achieves perplexity scores between DiLoCo-128 and DiLoCo-512 with $\approx4\times$ less communication than DiLoCo-512, $\approx16\times$ less than DiLoCo-128, and $\approx3000\times$ less than DDP. 
We demonstrate that similar trends hold for ImageNet-ResNet.

In our evaluation on a 4-node setup, all methods show a slight increase in perplexity compared to the 2-node case, yet the relative trends remain consistent. 
Our approach achieves perplexity scores between those of DiLoCo-128 and DiLoCo-512. 
These results suggest that our method is a promising, communication-efficient alternative for multi-node training in resource-constrained large-scale settings. 
However, a thorough investigation of its scaling properties across more nodes, more efficient communication mechanisms, and on higher-parameter models is left for future work.

\begin{table*}[h]
\centering
\resizebox{0.55\textwidth}{!}{%
\begin{tabular}{l|ccc|ccc|}
\cline{2-7}
& \multicolumn{3}{c|}{GPT-Neo-X (C4)} & \multicolumn{3}{c|}{ResNet (ImageNet-1k)} \\ \hline
\multicolumn{1}{|l|}{top-$k$} & \multicolumn{1}{c|}{32} & \multicolumn{1}{c|}{16} & 8    & \multicolumn{1}{c|}{32} & \multicolumn{1}{c|}{16}    & 8     \\ \hline \hline
\multicolumn{1}{|l|}{Perpl.}     & \multicolumn{1}{c|}{$36.3$} & \multicolumn{1}{c|}{$36.4$} & $35.6$ & \multicolumn{1}{c|}{$6.93$} & \multicolumn{1}{c|}{$11.42$} & $16.79$ \\ \hline
\multicolumn{1}{|l|}{Com. in GB} & \multicolumn{1}{c|}{$6.7$}  & \multicolumn{1}{c|}{$4.7$}  & $1.6$  & \multicolumn{1}{c|}{$1.9$}  & \multicolumn{1}{c|}{$0.9$} & $0.5$   \\ \hline
\end{tabular}
}
\caption{Perplexity and communication cost across models, datasets, and top-$k$ values}
\label{tab:perplexity_comm}
\end{table*}

\subsection{Analyzing Compression Intensity via the Top-$k$ Parameter}

Table~\ref{tab:perplexity_comm} shows the results over different top-$k$ parameters.
For the ResNet model, we observe the expected behavior: increasing $k$ leads to higher validation perplexity, as stronger compression is applied.
On the C4 dataset with a transformer model, however, the $k=8$ setting achieves the best perplexity, which is counter-intuitive.
Overall, perplexity remains relatively uniform across $k$.
We attribute the slight variations in perplexity observed in our transformer experiments to noise and suspect that specific properties of the momentum signal, arising from the loss surface of transformer language models, make them more robust to compression.
In future work, we aim to investigate these properties in more depth, with the goal of explaining them and identifying underlying principles that would allow us to control the degree of compression during training.
Such insights could benefit not only distributed model training but also improve training efficiency in centralized DDP settings.

\section{Conclusion}
Training large neural networks across multiple nodes remains resource-intensive. In this work, we propose effective combination of the strengths of federated learning local-global update strategy with momentum compression. We significantly reduce communication—by up to $16\times$ compared to DiLoCo and up to $3000\times$ compared to DDP.  Our results suggest that large-scale distributed training can become far more flexible—no longer tied to tightly coupled accelerators or expensive high-speed interconnects—if we rethink how and what we synchronize.
These findings highlight the potential for optimizing the communication efficiency of distributed model training and, we hope, will inspire further research into distributed training methods.

\bibliography{sample}
\newpage
\clearpage
\appendix

\section{Supplementary material}
\subsection{Distributed Low-Communication Training -- DiLoCo Algorithm Description}
\label{appendix:diloco}

\begin{algorithm2e}[H]
\caption{DiLoCo Algorithm}
\label{alg:diloco}
\SetKwInOut{Require}{Require}
\Require{Initial model $\theta^{(0)}$, $W$ workers, Data shards $\{D_1, \ldots, D_K\}$,  Optimizers \texttt{InnerOpt} and \texttt{OuterOpt}}

\For{$t \gets 1$ \KwTo $T$}{%
    \For{$i \gets 1$ \KwTo $W$}{%
        $\theta_i^{(t)} \gets \theta^{(t-1)}$\;
        \For{$h \gets 1$ \KwTo $H$}{%
            Sample $x \sim D_i$\;
            $L \gets f(x, \theta_i^{(t)})$\;
            \tcp{Inner optimization step}
            $\theta_i^{(t)} \gets \texttt{InnerOpt}(\theta_i^{(t)}, \nabla L)$\;
        }
    }
    \tcp{Averaging outer gradients}
    $\Delta^{(t)} \gets \frac{1}{W} \sum_{i=1}^{W} (\theta^{(t-1)} - \theta_i^{(t)})$\;
    \tcp{Outer optimization step}
    $\theta^{(t)} \gets \texttt{OuterOpt}(\theta^{(t-1)}, \Delta^{(t)})$\;
}
\end{algorithm2e}

\subsection{Decoupled Momentum Optimization: DeMo Algorithm Overview}
\label{appendix:demo}
\begin{algorithm2e}[H]
\caption{Decoupled Momentum Optimization}
\label{alg:decoupled_momentum}
\SetKwInOut{Require}{Require}

\Require{Learning rate $\eta$, decay $\beta \in (0, 1)$, parameters $\theta_t$, momentum $m_t$, number of high-frequency components $k$}

$\Delta L_t \gets \textsc{LocalStochasticGradient}(x_t)$ \\
$m_t \gets \beta m_t + \Delta L_t$ \\
$q_t \gets \textsc{ExtractHighFreqComponents}(m_t, k)$ \\
$m_{t+1} \gets m_t - q_t$ \\
$Q_t \gets \textsc{Synchronize}(q_t)$ \\
$\theta_{t+1} \gets \theta_t - \eta Q_t$ \\

\end{algorithm2e}

\subsection*{Discrete Cosine Transform: Rationale and Implementation}

We follow~\cite{peng2024demodecoupledmomentumoptimization} and use the DCT as a practical, decorrelating transform for energy compaction. While the optimal principal component decomposition (KLT) is intractable for large tensors, the DCT serves as a computationally efficient, highly parallelizable approximation. For signals with strong spatial correlation, the DCT closely approaches the KLT, and its fixed orthogonal basis enables exact decoding without auxiliary information.

Concretely, each momentum tensor $m$ (of shape $(n_0, n_1, ..., n_{d-1})$) is chunked along each axis into blocks of shape $(s_0, ..., s_{d-1})$ with $s_i | n_i$. We then apply a separable $d$-dimensional DCT to each chunk. The top-$k$ frequencies (by amplitude) are extracted and treated as principal components:
\begin{equation}
\tilde{m}_{\text{freq}}, \ \tilde{m}_{\text{ampl}} = p(m, s, k)
\end{equation}
After extracting these frequencies, the compressed representation comprises two tensors for each chunk: integer frequency indices and floating point amplitudes. Momentum reconstruction is performed via inverse DCT:
\begin{equation}
q_t = p^{-1}(\tilde{m}_{\text{freq}}, \ \tilde{m}_{\text{ampl}})
\end{equation}
All transform matrices are precomputed per chunk shape, so computational and memory overheads are minimal on modern hardware. Empirically, DCT compaction is sufficient to approximate the principal directions of momentum for efficient distributed optimization~\cite{peng2024demodecoupledmomentumoptimization}.

\subsection{Intuitive explanation on the convergence}\label{appendix:interaction}

At each synchronization step, we extract and synchronize the high-frequency components across all workers. Because these represent the most impactful, rapidly changing directions in the gradient, sharing them frequently accelerates convergence by aligning all workers on the most critical updates.

The low-frequency parts are kept locally as residuals and are not immediately synchronized. However, over multiple iterations, these components accumulate and can become more prominent—effectively. Accumulating low-frequency components allows mixing and transforming the accumulated results ones as the optimization landscape changes. We draw parallel to error-feedback SGD with momentum~\cite{karimireddy2019errorfeedbackfixessignsgd} where it is shown that when compressing gradients, but store residual error (delta), then optimization still converges. Because the algorithm continues extracting the top-$k$ components in subsequent iterations, eventually low-frequency components gain enough magnitude or importance to be captured in the high-frequency set. This ensures that no significant gradient information is permanently ignored; low-frequency information is progressively integrated and synchronized over time.
While this provides intuitive explanation, we leave theoretical proofs as future work.

\subsection{Additional details to the experimental setup}
\label{appendix:experimental_setup}
We provide further details about the experimental setup as follows. The GPT-NEO-X model comprises 3 hidden layers, each containing 16 self-attention heads, with an embedding dimension of 896, while the ResNet contains 50 residual blocks. 
For the language modeling task we used the C4 English dataset~\cite{raffel2023exploringlimitstransferlearning}, while for the image classification we used ImageNet-1k~\cite{5206848}. 
In our setup, we used 4 NVIDIA A100 GPUs. 
For all experiments, we used batch size of 512 with local gradient accumulation in case the full batch does not fit on the compute node. 
As reported in ~\cite{douillard2024dilocodistributedlowcommunicationtraining} we used optimal parameters for DiLoCo, and DDP, which we keep the same for our method. We explicitly point out specific hyperparameter changes in the experiment descriptions if we have any.
Due to computational constraints, we did not perform exhaustive hyperparameter optimization.

\end{document}